\documentclass[conference]{IEEEtran}

\IEEEoverridecommandlockouts
\usepackage{cite}
\usepackage{amsmath,amssymb,amsfonts}
\usepackage{graphicx}
\usepackage{textcomp}
\usepackage{xcolor}
\usepackage[ruled]{algorithm2e}
\usepackage{iftex}
\usepackage{enumitem}
\usepackage{subfigure}
\usepackage{multirow}

\makeatletter
\newcommand{\removelatexerror}{\let\@latex@error\@gobble}
\makeatother
\def\BibTeX{{\rm B\kern-.05em{\sc i\kern-.025em b}\kern-.08em
    T\kern-.1667em\lower.7ex\hbox{E}\kern-.125emX}}

\begin{document}

\title{ScaleDL: Towards Scalable and Efficient Runtime Prediction for Distributed Deep Learning Workloads}

\author{
    Xiaokai Wang$^{\dag}$, Shaoyuan Huang$^{\dag}$, Yuting Li$^{\dag}$, and Xiaofei Wang$^{\dag}$\\
    \IEEEauthorblockA{$^{\dag}$College of Intelligence and Computing, Tianjin University, Tianjin, China} 
    Email: \{xiaokaiwang, hsy\_23, 2020214010, xiaofeiwang\}@tju.edu.cn
}

\maketitle

\begin{abstract}
Deep neural networks (DNNs) form the cornerstone of modern AI services, supporting a wide range of applications, including autonomous driving, chatbots, and recommendation systems.
As models increase in size and complexity, DNN workloads such as training and inference tasks impose unprecedented demands on distributed computing resources, making accurate runtime prediction essential for optimizing development and resource allocation.
Traditional methods rely on additive computational unit models, limiting their accuracy and generalizability. In contrast, graph-enhanced modeling improves performance but significantly increases data collection costs. Therefore, there is a critical need for a method that strikes a balance between accuracy, generalizability, and data collection costs.
To address these challenges, we propose ScaleDL, a novel runtime prediction framework that combines nonlinear layer-wise modeling with graph neural network (GNN)-based cross-layer interaction mechanism, enabling accurate DNN runtime prediction and hierarchical generalizability across different network architectures. Additionally, we employ the D-optimal method to reduce data collection costs.
Experiments on the workloads of five popular DNN models demonstrate that ScaleDL enhances runtime prediction accuracy and generalizability, achieving 6$\times$ lower MRE and 5$\times$ lower RMSE compared to baseline models.
\end{abstract}


\section{Introduction}

DNNs have become essential to modern AI services, contributing to advancements in fields such as recommendation systems~\cite{rec_survey}, computer vision~\cite{segment}, and natural language processing~\cite{llama}. 
With the rapid expansion of model scale and application diversity, DNN training and inference generate massive, heterogeneous workloads, imposing unprecedented demands on distributed computing resources such as computation, memory, and communication bandwidth.
Meanwhile, deployment environments have become increasingly diverse, including GPUs, TPUs, and custom accelerators—each exhibiting distinct performance characteristics\cite{Habitat}.
Against this backdrop of ever-growing model scale, heterogeneous accelerators, and software stacks, achieving a precise understanding of DNN workload runtime performance—including \textit{computation time} and \textit{communication overhead}—has become even more critical~\cite{hpc_survey, oneforall}.

Runtime prediction has emerged as a key enabler for both system-level optimization and model development in large-scale DNN workloads~\cite{distDNN_survey}.
At the system level, reliable predictions enable model-device co-optimization, dynamic scaling in cloud environments, and efficient scheduling in distributed training~\cite{258862, seer}. In real-world development, they help mitigate risks by preventing job failures and excessive resource consumption, enabling practitioners to identify feasible configurations, manage budgets, and accelerate experimentation.







Recent research in DNN workload runtime prediction has witnessed significant strides ~\cite{Habitat,paleo,MICRO23,ml_performance,DNNPerf,DeepPerf,PerfSeer}.
Traditional approaches typically assume a linear relationship with the number of floating-point operations (FLOPs)~\cite{paleo}, providing coarse estimates based on computational complexity but failing to capture nonlinear effects, such as memory bottlenecks and activation functions, which leads to suboptimal accuracy. 
Operator-level modeling~\cite{MICRO23,ml_performance,Habitat} improves on this by modeling DNNs at a finer operator granularity and capturing additional nonlinearities within operators, but it still overlooks interactions between operators.
More recently, graph-based methods, such as those employing GNNs\cite{DeepPerf,DNNPerf}, have shown advantages in capturing model structural dependencies. However, these purely neural network approaches introduce high data collection costs and limited generalizability across different models.
Therefore, there is an urgent need for a runtime prediction framework that strikes a balance between the three objectives illustrated in Figure~\ref{fig:challenge}:

\begin{figure}
    \centering
    \includegraphics[width=1.0\linewidth]{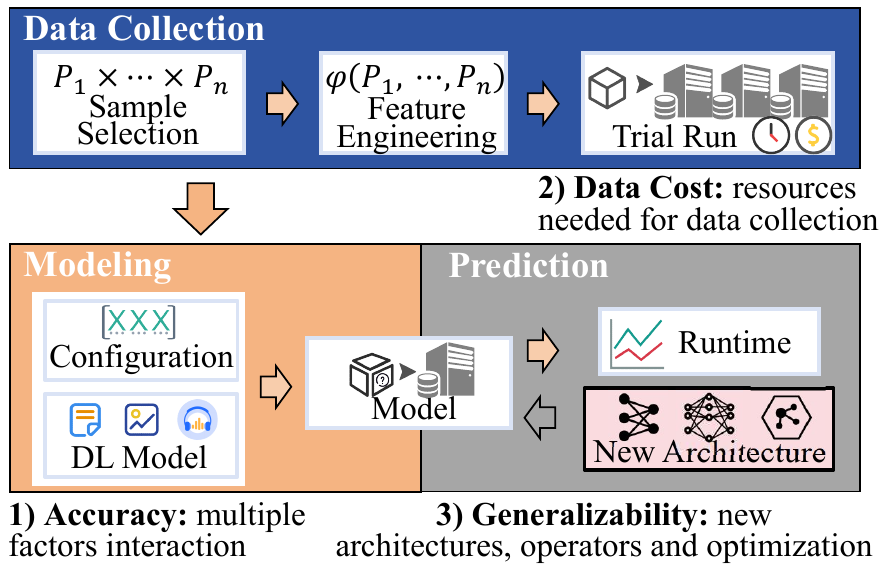}
    \caption{Objectives of DNN workloads runtime prediction.}
    \label{fig:challenge}
    \vspace{-0.5cm}
\end{figure}

\begin{enumerate}{}
    \item \emph{\textbf{Accuracy}}. The runtime performance of DNNs is jointly determined by multiple factors, including primary computational layers, configuration parameters (e.g., batch size, input/output dimensions and parallelism), hardware platforms (e.g., GPU, TPU), software frameworks (e.g., TensorFlow, PyTorch), and system-level optimizations (e.g., kernel fusion, memory management). The complex and nonlinear interactions among these components make it challenging to accurately capture runtime behaviors, often resulting in suboptimal prediction accuracy.
    \item \emph{\textbf{Data Cost}}. The above factors form a vast configuration space, making exhaustive profiling of configurations impractical. For instance, benchmarking a single ViT training configuration can take several GPU hours, and enumerating all possible parameter combinations would require thousands more. Hence, an effective approach must balance data efficiency with predictive accuracy.
    \item \emph{\textbf{Generalizability}}. As DNN architectures rapidly evolve and new operators and optimizations emerge, static or model-specific predictors quickly become obsolete. Therefore, a robust prediction framework should generalize to unseen models with minimal re-benchmarking, ensuring scalability and practicality in real deployments.
\end{enumerate}


To address these challenges, we propose ScaleDL, a scalable and efficient runtime prediction framework for distributed DNN workloads. 
Specifically, ScaleDL incorporates the following three key designs:
\textbf{1) layer-wise modeling.}
Instead of modeling the entire network as a monolithic entity, ScaleDL decomposes DNNs into individual layers and predicts their execution performance separately. 
This design significantly reduces data collection requirements while providing high generalizability;
\textbf{2) cross-layer interaction modeling.}
A lightweight GNN is employed to capture inter-layer dependencies such as computation overlap, kernel fusion, and memory reuse—factors that traditional linear or operator-level models fail to represent;
\textbf{3) data-efficient training.}
ScaleDL employs a D-optimal sampling strategy to train the model with only a subset of informative configurations, striking a balance between profiling overhead and accuracy. Our contributions are summarized as follows:

\begin{itemize}
\item We propose ScaleDL, a scalable runtime prediction framework that integrates fine-grained layer-wise modeling with GNN-based cross-layer interaction to accurately predict distributed DNN workloads.
\item Leveraging layer-wise modeling and a D-optimal sampling strategy, ScaleDL reduces the required benchmarking samples and overall collection cost by up to 5.2$\times$ without compromising prediction accuracy.
\item Experiments on the workloads of five popular DNN models show that ScaleDL achieves a 6$\times$ reduction in MRE and 5$\times$ in RMSE compared to baseline frameworks.
\end{itemize}

\section{ScaleDL System Overview}

\subsection{ScaleDL Overview}
\begin{figure}[tbp]
\centerline{\includegraphics[width=\linewidth]{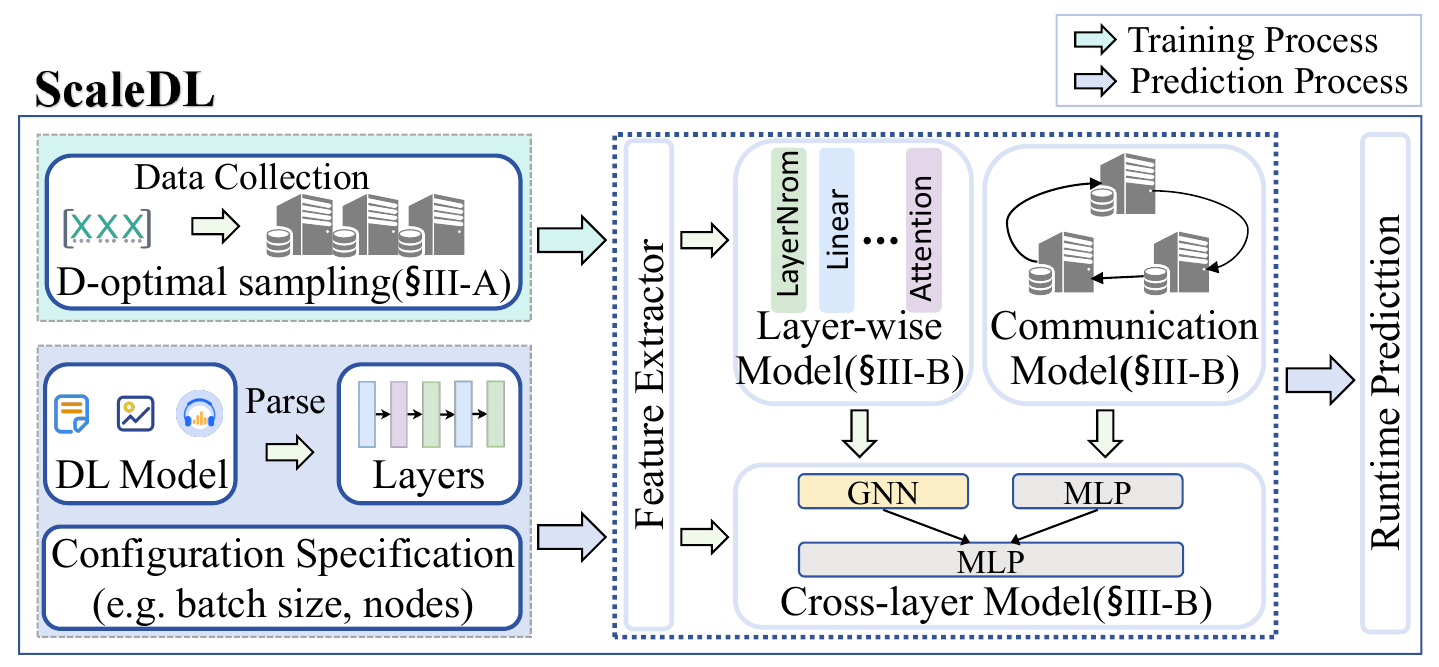}}
\caption{Overview of ScaleDL.}
\label{fig: ScaleDL}
\vspace{-0.4cm}
\end{figure}
Figure~\ref{fig: ScaleDL} shows the overall architecture of ScaleDL. 
It first predicts the runtime of DNN workloads by modeling the workloads as a combination of computational cost and communication delay.
ScaleDL then uses graph correction to capture cross-layer effects, such as kernel fusion and data dependencies, refining the initial estimates provided by the layer-wise cost and communication performance models.
To further achieve data-efficient training, ScaleDL employs a D-optimal sampling strategy. By optimizing the required data volume, it ensures the model efficiently learns patterns that provide the most informative value for runtime prediction.

\subsection{Runtime Model} \label{sec:graph}
We define the runtime for a single DL training epoch as $T_\text{epoch}$, which is primarily determined by the time spent on its iterations. Under steady-state execution, excluding warm-up activities such as data loading and container startup, the per-iteration time remains approximately constant. Assuming $I$ is the number of iterations per epoch, we model the per-iteration compute latency $T_\text{iter}$ and compute the total epoch time as:
\begin{gather}\label{eq:Tepoch}
    T_{\text{epoch}} = I\cdot T_{\text{iter}}
\end{gather}
and we model $T_{\text{iter}}$ as:
\begin{gather}\label{eq:Titer}
    T_\mathrm{iter} = \alpha \cdot T_{\mathrm{sum}}  + \beta \cdot T_\mathrm{comm} 
\end{gather}
where $T_{\text{sum}}$ represents the DNN computation time, and $T_\mathrm{comm}$ is the communication time, the modeling of which will be elaborated in the next section.
To model the interaction between computation and communication, we represent one iteration of a DNN as a directed acyclic graph $G = (V, E)$, where each node $v \in V$ corresponds to a layer and edges $E$ capture dependencies between layers.
This graph-based architecture models how each layer’s computation depends on both preceding and succeeding layers,
and is used to model the computation time scaling factor $\alpha$ and the communication time factor $\beta$ as follows:
\begin{gather}\label{eq:gnn}
    \{ \alpha, \beta \} = \Omega\left( G; \{ \mathbf{v}_i \}_{i \in V}, \{ \mathbf{e}_{ij} \}_{(i,j) \in E}, \mathcal{Z} \right)
\end{gather}
where:
\begin{itemize}[leftmargin=*, itemsep=0.01pt]
\item  $\Omega$ indicates a GNN-based model that predicts scaling factors based on input graph information $G$.
 \item $\{\mathbf{v}_i\}_{i \in V}$ are node features, capturing key layer attributes, e.g., type, FLOPs, and predicted execution time. 
 \item $\{\mathbf{e}_{ij}\}_{(i,j) \in E}$ are edge features, encoding dependencies between layers, such as data transfer size and direction.
 \item $\mathcal{Z}$ represents global hyperparameters, such as batch size, optimizer type, and hardware characteristics.
\end{itemize}

\section{Methodology}
\subsection{Efficient D-optimal Sampling}

In the data collection stage, the objective is to acquire informative performance samples within a limited budget for training both the layer-wise predictors and the graph-based model. Due to the high-dimensional configuration space and GPU-intensive benchmarks, exhaustive or uniform sampling becomes inefficient with inflated costs. To address this, we employ a D-optimal experimental design for our modeling pipeline, selecting as few samples as possible while maximizing diversity and identifiability in the feature space. Assume each candidate configuration is represented as a feature vector $a_i$, where $i\in[1,m]$, and we define the following:
\begin{equation}
    M(\lambda) = \sum_{i=1}^m\lambda_ia_ia_i^\top 
\end{equation}
which represents the empirical feature information matrix, where $\lambda_i \in \{0,1\}$ indicates whether $a_i$ is selected. 
Maximizing the determinant $\mathrm{det} (M(\lambda))$ enhances the information about the model parameters, thereby minimizing the variance of their estimates. In our nonlinear settings, this criterion minimizes uncertainty in the feature covariance, ensuring selected samples span the space as uniformly and orthogonally as possible. Given a fixed sample budget $k$, we solve a combinatorial D-optimal subset selection problem:
\begin{align}
&\max_{\boldsymbol{\lambda}}\;\;  \mathrm{det}(M(\lambda)) \\
\text{s.t.}\;\; & \sum_{i=1}^m \lambda_i = k, \lambda_i \in \{0,1\}. \notag
\end{align}

To solve this, we apply the Fedorov-exchange heuristic and benchmark only the configurations where $\lambda_i = 1$. This ensures both the layer-level regressors $\phi_{l}(\cdot)$ and the graph-level model $\Omega(\cdot)$ are trained on highly informative, non-redundant configurations. By leveraging D-optimal techniques and multi-process concurrency optimization, the collection of layer benchmarks and GNN training data in ScaleDL is completed within two hours, ensuring data-efficient training.

\subsection{DNN Runtime Model}

\begin{figure}
    \centering
    \includegraphics[width=1.0\linewidth]{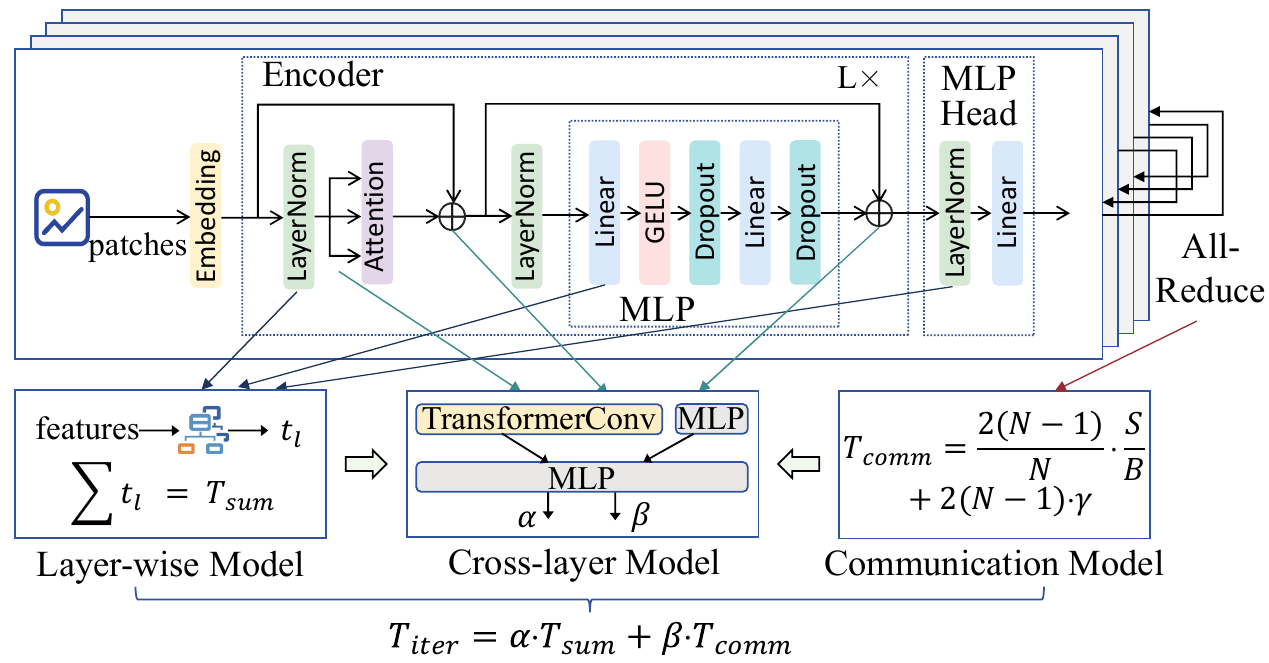}
    \caption{DNN runtime model within ScaleDL for ViT.}
    \label{fig:vit}
    \vspace{-0.4cm}
\end{figure}

Accurate runtime prediction requires modeling layer behavior, inter-layer dependencies, and the interaction between computation and communication. As shown in Figure~\ref{fig:vit}, we propose a three-stage runtime model within ScaleDL: 
(i) Layer-wise modeling, which analyzes the characteristics and computational demands of each layer type and uses type-specific per-layer predictors to generate initial computation time estimates; 
(ii) cross-layer interaction modeling, in which a graph-level model refines these estimates by capturing cross-layer effects such as kernel fusion, memory reuse, and the coupling between computation and communication; 
and (iii) communication performance modeling, which leverages interpretable formulas to model the widely used all-reduce pattern to estimate communication overhead.

\textbf{Layer-wise modeling}. \label{sec:layer-model}
We predict the computation cost of each layer and accumulate these latencies to estimate the total computation latency for one iteration, denoted as:
\begin{gather}
    T_{\text{sum}} = \sum_{l \in V} \mathbf{t}_l
\end{gather}
where $\mathbf{t}_l$ represents the computation cost of layer $l$.

To predict the computation cost of each layer, we develop a separate Random Forest regression predictor for each layer, accounting for relevant features such as hyperparameters, device characteristics, and layer configurations. 

For each layer $l$, the computation cost $t_l$ is predicted as:
\begin{gather}\label{eq:tl}
    \mathbf{t}_l \;=\; \phi_{l}\!\big(\mathbf{h}^{\mathrm{hp}} \,\Vert\, \mathbf{h}^{\mathrm{dev}} \,\Vert\, \mathbf{h}^{\mathrm{struct}} \,\Vert\, \mathbf{h}^{\mathrm{cm}}\big)
\end{gather}
where:
\begin{itemize}
    \item $\mathbf{h}^{\mathrm{hp}}$: Core hyperparameters that influence the model’s performance, such as batch size and sequence length.
    \item $\mathbf{h}^{\mathrm{dev}}$: Device-specific characteristics that define the hardware's computational capabilities, including peak compute performance and memory bandwidth.
    \item $\mathbf{h}^{\mathrm{struct}}$: Layer-specific configuration features that characterize the layer's architecture and design.
    \item $\mathbf{h}^{\mathrm{cm}}$: Computational and memory-related metrics, such as FLOPs (floating-point operations) and tensor byte count, that capture the resource usage of the layer.
\end{itemize}


To better understand $\mathbf{h}^{\mathrm{struct}}$ and $\mathbf{h}^{\mathrm{cm}}$, we take the linear layer as an example. The structural features $\mathbf{h}^{\mathrm{struct}}$ include the input dimension $d_{\text{in}}$ and the output dimension $d_{\text{out}}$. The characteristics of $\mathbf{h}^{\mathrm{cm}}$ are defined as:
\begin{align}
    \mathbf{h}^{\mathrm{cm}}_{\text{Linear}} = (\text{FL}&\text{OPs}_\text{Linear}, \text{Params}_\text{Linear}) \\
    \text{FLOPs}_\text{Linear} &= 2b d_{\text{in}} d_{\text{out}} + b d_{\text{out}} \\
    \text{Params}_\text{Linear} &= d_{\text{in}} d_{\text{out}} + d_{\text{out}}
\end{align}

where $b$ denotes the batch size. 
$\text{FLOPs}_\text{Linear}$ include the floating-point operations for multiply-accumulate and bias addition in the Linear layer. $\text{Params}_\text{Linear}$ represent memory usage based on the number of parameters.
This approach lays the groundwork for end-to-end runtime prediction, estimating the computation latency, $T_{\text{sum}}$, which is further refined by the GNN model through capturing layer dependencies.

\addtolength{\topmargin}{0.01in}
\begin{algorithm}[t]
    \caption{\emph{ScaleDL} Prediction Process}
    \label{alg:scaleDL_process}
    \SetAlgoNlRelativeSize{-3}
    \SetKwInOut{Input}{Input}
    \SetKwInOut{Output}{Output}

    \Input{
        Model Architecture \( \mathcal{M} \), Configuration \( \mathcal{C} = (\text{hyperparameters}, \text{hardware features}, N, B, \gamma) \).
    }
    \Output{
        Predicted epoch execution time \( T_{\text{pred}} \).
    }

    \vspace{0.1cm}
    \textbf{Stage 1: Compute Time Prediction} \\
    \(\mathcal{L} \gets \text{Decompose}(\mathcal{M}) \) \\
    \For{each layer \( l \in \mathcal{L} \)}{
        Define \( \mathbf{h} = \left( \mathbf{h}^{\mathrm{hp}} \Vert \mathbf{h}^{\mathrm{dev}} \Vert \mathbf{h}^{\mathrm{struct}} \Vert \mathbf{h}^{\mathrm{cm}} \right) \) \\
        \( \mathbf{h} \gets \text{LayerFeatureExtractor}(l, \mathcal{C}) \) \\
        \( \mathbf{t}_{l} = \phi_{l}(\mathbf{h}) \)
    } 
    \(T_{\text{sum}} = \sum t_i \)

    \vspace{0.25cm} 
    \textbf{Stage 2: Communication Time Prediction} \\
    \( T_{\text{comm}} = \frac{2(N-1)}{N} \cdot \frac{S}{B} + 2(N-1) \cdot \gamma \)

    \vspace{0.25cm} 
    \textbf{Stage 3: Dual Refinement using GNN} \\
    Define \( \mathcal{V} = \{ \mathbf{v}_i \}_{i \in \mathcal{V}} \) and \( \mathcal{E} = \{ \mathbf{e}_{ij} \}_{(i,j) \in \mathcal{E}} \). \\
    \vspace{0.03cm}
    \(\{ \mathcal{V}, \mathcal{E}, \mathcal{Z} \} \gets \text{GNNFeatureExtractor}(\mathcal{L}, \mathcal{C}, \mathbf{t}_l) \) \\
    \( \{ \alpha, \beta \} = \Omega\left( G; \mathcal{V}, \mathcal{E}, \mathcal{Z} \right) \)

    \vspace{0.25cm} 
    \textbf{Final Step: Compute Predicted Execution Time} \\
    \( T_{\text{iter}} = \alpha \cdot T_{\text{sum}} + \beta \cdot T_{\text{comm}} \) \\
    \( T_{\text{pred}} = I \cdot T_{\text{iter}} \)

    \vspace{0.1cm}
    \Return \( T_{\text{pred}} \)
\end{algorithm}

\textbf{Cross-layer interaction modeling}.
To model the intricate interplay between computation and communication across layers, we employ a novel Transformer-based GNN to capture complex inter-node relationships.
The GNN incorporates a TransformerConv layer, which leverages the DNN’s computational flow graph to dynamically learn feature representations on nodes (DNN layers) and edges (cross-layer interactions).
At the core of this layer, an attention coefficient, $\epsilon_{i,j}$, is computed for each node pair by explicitly injecting edge-feature information into the attention score. This unified coefficient can be expressed as:
\begin{equation}
\epsilon_{i,j}=\frac{(W_Q v_i)^T (W_K v_j)}{\sqrt{d_k}}+\mathrm{ReLU}(W_{\mathrm{edge}} e_{i,j}+b_{\mathrm{edge}})
\end{equation}
where the two terms correspond to scaled dot-product attention on node features ($v_i, v_j$), and a learnable transformation of edge features ($e_{i,j}$), enabling the model to reason about structural dependencies. These raw scores are then normalized to a probability distribution, $\omega_{i,j}$, via the softmax function:
\begin{equation}
\omega_{i,j}=\mathrm{softmax}_j(\epsilon_{i,j})=\frac{\exp(\epsilon_{i,j})}{\sum_{k\in N(i)}\exp(\epsilon_{i,k})}
\end{equation}

To capture a rich spectrum of relational patterns, we employ a multi-head attention mechanism. The updated node representation, $v_i^{\prime}$, is obtained by averaging the outputs from $L$ parallel attention heads:
\begin{equation}
v_i^{\prime} = \frac{1}{L} \sum_{l=1}^{L} \sigma\left(\sum_{j\in N(i)}\omega_{i,j}^{(l)} W_V^{(l)} v_j\right)
\end{equation}
where $\omega_{i,j}^{(l)}$ and $W_V^{(l)}$ are the attention weights and the value transformation matrix for head $l$.

In addition to the TransformerConv layer, the GNN incorporates a global encoder and a predictor through an MLP layer. The encoder processes the global hyperparameters $\mathcal{Z}$, while the TransformerConv layer aggregates the node representations $v_i^{\prime}$ into a graph-level embedding, together forming a dual-branch architecture. The resulting vectors are concatenated and passed through the final predictor to generate the scaling factors $\alpha$ and $\beta$. Mathematically, the process is given by:
\begin{equation}
\{\alpha, \beta\}=\mathrm{MLP}\left(\left[\mathrm{Pooling}({v_i^{\prime}},{i=1,\dots,n})|\mathrm{MLP(\mathcal{Z})}\right]\right)
\end{equation}

This dual-branch approach, which synergistically combines the GNN's graph-based learning with the MLP's modeling of global information, proves highly effective in enhancing prediction accuracy, even with limited training data.




\textbf{Communication performance modeling.} To model communication costs, we focus on the all-reduce communication pattern, which is widely used in various distributed training frameworks.
Communication costs consist of two components: data transfer time and communication latency. Let $N$ represent the number of GPUs involved, $S$ the data size (in bits), $B$ the network bandwidth (in bits per second), and $\gamma$ the communication latency (in seconds). Using these parameters, the communication cost can be modeled as:
\begin{gather}\label{eq:comm}
    T_{\mathrm{comm}}=\frac {2(N-1)}{N}\cdot\frac SB+2(N-1)\cdot\gamma 
\end{gather}
where the first term, $\frac{2(N - 1)}{N} \cdot \frac{S}{B}$, represents the data transfer time. In the all-reduce operation, data is exchanged between all nodes, where $\frac{S}{B}$ is the time required to transfer one unit of data, and $\frac{2(N - 1)}{N}$ accounts for the bidirectional communication between multiple nodes. The second term, $2(N - 1) \cdot \gamma$, represents the communication delay. Since all-reduce is a synchronous operation, each node must wait for the others to complete their computations, introducing an additional delay $\gamma$, with $2(N - 1)$ reflecting the bidirectional latency.

\subsection{Runtime Prediction and Evaluation}
Before using the ScaleDL framework, the target DNN model and its configuration must be specified.
As detailed in Algorithm~\ref{alg:scaleDL_process}, the model is then decomposed into layers from which structural and computational features are extracted.
These features, along with the configuration parameters, form a unified feature vector $\mathbf{h}$, which is input into a pre-trained layer predictor to estimate execution times and summed to compute the baseline computation time, $T_\text{sum}$.

Subsequently, the communication time $T_{\text{comm}}$ is computed using theoretical formulas (eq.~\ref{eq:comm}), with the communication latency factor $\gamma$ measured through an all-reduce test. After $T_{\text{sum}}$ and $T_{\text{comm}}$ are obtained, we proceed by extracting the relevant edge features, node features, and hyperparameters. These are then input into the trained GNN model, which outputs the scaling factors $\alpha$ and $\beta$. Finally, using these scaling factors in conjunction with formulas (eq.~\ref{eq:Tepoch} and eq.~\ref{eq:Titer}), we calculate the final predicted execution time $T_{\text{pred}}$.

To evaluate model accuracy, we use two metrics: Mean Relative Error (MRE) and Root Mean Squared Error (RMSE). The MRE is defined as $ \text{MRE} = \frac{1}{n} \sum_{i=1}^{n} \frac{\left| T_{\text{epoch}}^i - T_{\text{pred}}^i \right|}{T_{\text{epoch}}^i}\times100\%$, which measures the relative difference between the predicted and actual values. The RMSE is given by $\text{RMSE} = \sqrt{\frac{1}{n} \sum_{i=1}^{n} \left( T_{\text{epoch}}^i - T_{\text{pred}}^i \right)^2}$, which captures the magnitude of errors. 
Here, $n$ represents the total number of samples, with $T_{\text{epoch}}^i$ denoting the actual training time for sample $i$, and $T_{\text{pred}}^i$ representing the predicted runtime. Both metrics indicate better performance with lower values.


\section{Experiment}
\subsection{Experiment Setup}
The experiment was conducted on two servers that were equipped with Intel Xeon E5-2690V4 CPUs and NVIDIA H20 GPUs. We benchmarked five representative DNNs: T5 \cite{T5}, GPT-2 \cite{gpt-2}, BERT \cite{Bert}, ViT \cite{ViT}, and DeiT \cite{DeiT}. 
Our analysis focuses on $T_\text{epoch}$ under DNN training tasks, as these workloads are computationally complex and encompass the entire inference computation process. Using the D-optimal method, we collected 200 distinct samples by varying different configurations for each DNN model. The dataset was then partitioned into 160 training samples and 40 testing samples. 

We evaluated the accuracy and generalizability of ScaleDL in the In-Domain (ID) and Out-of-Domain (OOD) settings\cite{RooflineLLM}, calculating MRE and RMSE. 
The ID scenario assesses the model's ability to fit by training and testing on the same DNN architecture. In contrast, the OOD scenario evaluates generalization by training on a set of DNNs excluding the target architecture, testing on unseen architectures.


For comparison, we selected the following three baseline frameworks: 1) Random Forest (RF), a simple regression model using only hyperparameters, without considering the graph structure; 2) MLP-ACC \cite{ml_performance}, an MLP-based model that sums individual layer time predictions; and 3) BiRNN \cite{BiRNN}, a bidirectional RNN that models sequential dependencies using node features but lacks edge-feature information.


\subsection{Accuracy of Runtime Predictions}
\begin{table}[]
\caption{Accuracy in Predicting Runtime Across Different Models (ID).}
\label{table:ID_result}
\begin{tabular}{|c|ccccc|}
\hline
\multirow{2}{*}{\textbf{\begin{tabular}[c]{@{}c@{}}Model\\ Name\end{tabular}}} & \multicolumn{5}{c|}{\textbf{Prediction of Training Time}}                                                                                                                                                                                                                                                                                                                              \\ \cline{2-6} 
                                                                               & \multicolumn{1}{c|}{\textbf{Metrics}}                                           & \multicolumn{1}{c|}{\textbf{ScaleDL}}                                               & \multicolumn{1}{c|}{\textbf{RF}}                                            & \multicolumn{1}{c|}{\textbf{MLP-ACC}}                                       & \textbf{BiRNN}                                         \\ \hline
Overall                                                                        & \multicolumn{1}{c|}{\begin{tabular}[c]{@{}c@{}}MRE(\%)\\ RMSE(ms)\end{tabular}} & \multicolumn{1}{c|}{\textbf{\begin{tabular}[c]{@{}c@{}}9.39\\ 43.66\end{tabular}}}  & \multicolumn{1}{c|}{\begin{tabular}[c]{@{}c@{}}35.75\\ 170.89\end{tabular}} & \multicolumn{1}{c|}{\begin{tabular}[c]{@{}c@{}}15.81\\ 69.77\end{tabular}}  & \begin{tabular}[c]{@{}c@{}}19.78\\ 88.41\end{tabular}  \\ \hline
T5                                                                             & \multicolumn{1}{c|}{\begin{tabular}[c]{@{}c@{}}MRE(\%)\\ RMSE(ms)\end{tabular}} & \multicolumn{1}{c|}{\textbf{\begin{tabular}[c]{@{}c@{}}9.45\\ 24.39\end{tabular}}}  & \multicolumn{1}{c|}{\begin{tabular}[c]{@{}c@{}}33.84\\ 97.55\end{tabular}}  & \multicolumn{1}{c|}{\begin{tabular}[c]{@{}c@{}}14.57\\ 37.48\end{tabular}}  & \begin{tabular}[c]{@{}c@{}}20.15\\ 64.71\end{tabular}  \\ \hline
GPT-2                                                                          & \multicolumn{1}{c|}{\begin{tabular}[c]{@{}c@{}}MRE(\%)\\ RMSE(ms)\end{tabular}} & \multicolumn{1}{c|}{\begin{tabular}[c]{@{}c@{}}\textbf{13.88}\\ 118.13\end{tabular}}         & \multicolumn{1}{c|}{\begin{tabular}[c]{@{}c@{}}42.78\\ 155.81\end{tabular}} & \multicolumn{1}{c|}{\begin{tabular}[c]{@{}c@{}}18.68\\ \textbf{46.51}\end{tabular}}  & \begin{tabular}[c]{@{}c@{}}23.47\\ 150.84\end{tabular} \\ \hline
Bert                                                                           & \multicolumn{1}{c|}{\begin{tabular}[c]{@{}c@{}}MRE(\%)\\ RMSE(ms)\end{tabular}} & \multicolumn{1}{c|}{\textbf{\begin{tabular}[c]{@{}c@{}}5.47\\ 55.16\end{tabular}}}  & \multicolumn{1}{c|}{\begin{tabular}[c]{@{}c@{}}34.57\\ 509.60\end{tabular}} & \multicolumn{1}{c|}{\begin{tabular}[c]{@{}c@{}}12.42\\ 106.69\end{tabular}} & \begin{tabular}[c]{@{}c@{}}21.13\\ 165.89\end{tabular} \\ \hline
ViT                                                                            & \multicolumn{1}{c|}{\begin{tabular}[c]{@{}c@{}}MRE(\%)\\ RMSE(ms)\end{tabular}} & \multicolumn{1}{c|}{\textbf{\begin{tabular}[c]{@{}c@{}}7.74\\ 9.63\end{tabular}}}   & \multicolumn{1}{c|}{\begin{tabular}[c]{@{}c@{}}23.95\\ 48.44\end{tabular}}  & \multicolumn{1}{c|}{\begin{tabular}[c]{@{}c@{}}15.89\\ 29.07\end{tabular}}  & \begin{tabular}[c]{@{}c@{}}16.69\\ 37.77\end{tabular}  \\ \hline
DeiT                                                                           & \multicolumn{1}{c|}{\begin{tabular}[c]{@{}c@{}}MRE(\%)\\ RMSE(ms)\end{tabular}} & \multicolumn{1}{c|}{\textbf{\begin{tabular}[c]{@{}c@{}}10.39\\ 10.99\end{tabular}}} & \multicolumn{1}{c|}{\begin{tabular}[c]{@{}c@{}}43.63\\ 43.04\end{tabular}}  & \multicolumn{1}{c|}{\begin{tabular}[c]{@{}c@{}}17.47\\ 129.12\end{tabular}} & \begin{tabular}[c]{@{}c@{}}17.44\\ 22.84\end{tabular}  \\ \hline
\end{tabular}
\end{table}

\begin{figure}[tbp]
\centerline{\includegraphics[width=1.0\linewidth]{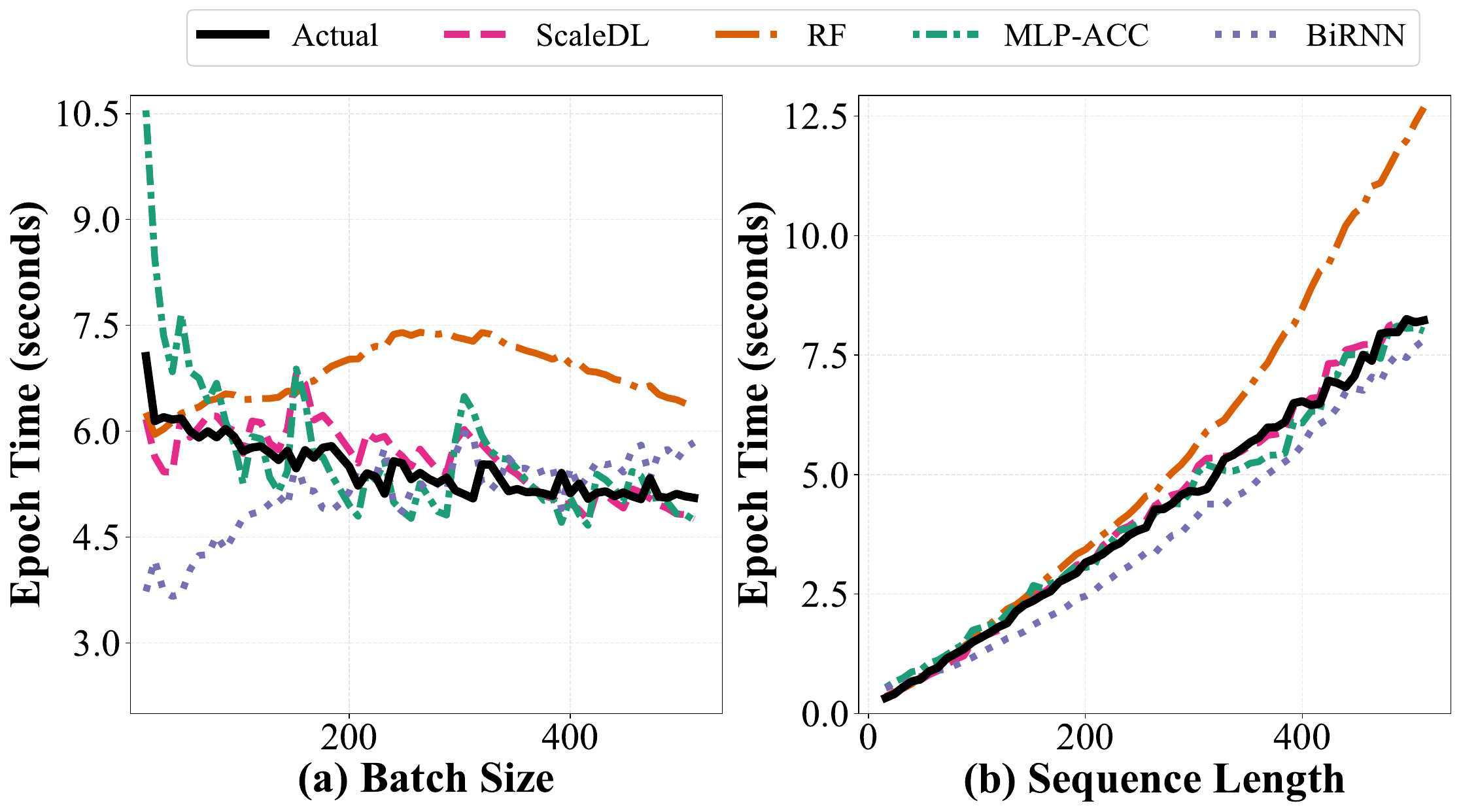}}
\caption{Accuracy in predicting runtime for the BERT model across different key parameters on a fixed dataset: (a) epoch runtime vs. batch size and (b) epoch runtime vs. sequence length.}
\label{fig:combined}
\vspace{-0.4cm}
\end{figure}

As shown in Table~\ref{table:ID_result}, ScaleDL outperforms all baseline methods in predicting training time across various workloads, with an overall MRE of 9.39\% and RMSE of 43.66 ms.
Specifically, it reduces the MRE and RMSE by 73.7\% and 74.5\% compared with RF, by 40.6\% and 37.4\% compared with MLP-ACC, and by 52.5\% and 50.6\% compared with BiRNN, respectively.
For specific DNN models like T5, Bert, and ViT, ScaleDL consistently delivers superior performance, with the best MRE of 5.47\% for Bert and the lowest RMSE of 9.63 ms for ViT. While MLP-ACC slightly outperforms ScaleDL in RMSE for GPT-2, ScaleDL still leads in MRE. This demonstrates ScaleDL’s ability to capture cross-layer interactions, resulting in higher accuracy than the other models.

Figure~\ref{fig:combined} further illustrates ScaleDL's ability to capture execution time trends with varying DNN configuration parameters. As batch size increases, the total execution time for an epoch decreases due to enhanced parallelism. RF fails to capture this trend, while BiRNN and MLP-ACC perform slightly better but still struggle to predict execution time accurately, especially at smaller batch sizes. 
For sequence length, execution time increases linearly, with all models showing a similar trend. However, RF and BiRNN exhibit larger deviations at higher sequence lengths. While MLP-ACC performs better, it still lags behind ScaleDL. In contrast, ScaleDL accurately tracks changes in execution time, demonstrating its superior ability to model DNN runtime as parameters vary.



\subsection{Generalizability of Runtime Predictions}
\begin{figure}[tbp]
\centerline{\includegraphics[width=0.8\linewidth]{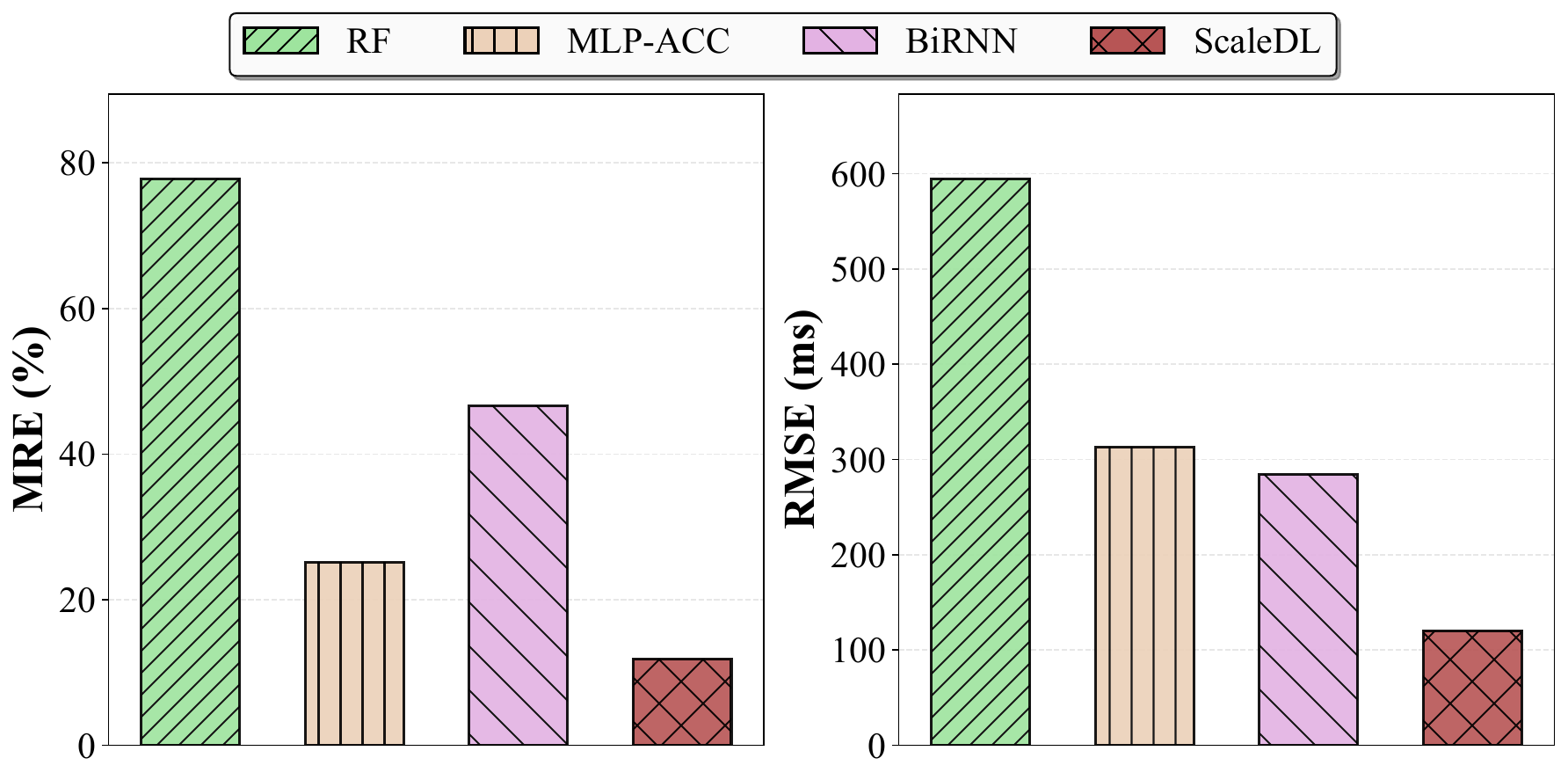}}
\caption{Average accuracy across different DNN models under OOD settings. Prediction frameworks are trained on all data without the target DNN.}
\label{fig:ood}
\vspace{-0.3cm}
\end{figure}

In the OOD scenario, we performed a generalization test for each target model and then computed the average prediction metrics across all models to assess generalizability.
As shown in Figure~\ref{fig:ood}, ScaleDL significantly outperforms all baselines, with a wider margin than in the ID scenario. Specifically, ScaleDL’s MRE is 11.88\%, approximately 6$\times$ smaller than RF at 77.81\%, and its RMSE is 120.5 ms, about 5$\times$ lower than RF at 594.7 ms. For MLP-ACC, the MRE is 25.14\%, and the RMSE is 313.1 ms, both at least 2$\times$ higher than ScaleDL’s results. Similarly, BiRNN shows 46.65\% MRE and 284.3 ms RMSE, with 2.4$\times$ higher than ScaleDL's. These results highlight ScaleDL's strong generalizability, which stems from its layer-wise and cross-layer interaction modeling structure.



\subsection{Ablation Study}

\begin{figure}
\centering
\subfigure[Training samples under different sampling strategies.]{\label{fig:samples}
\includegraphics[width=2.6cm]{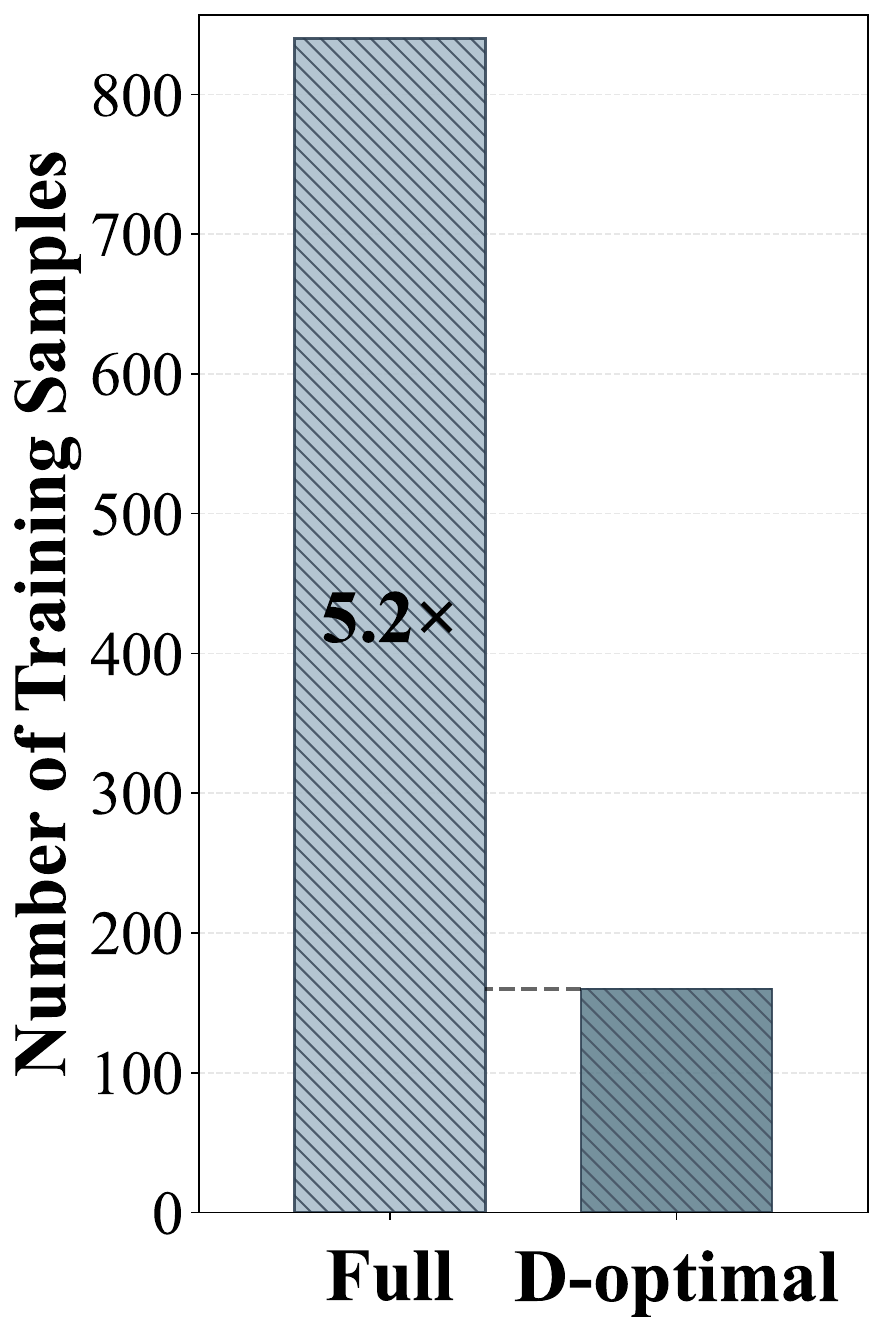}
}
\subfigure[The Cumulative Distribution Function (CDF) of absolute percentage error for each sample, with MRE marked for differences frameworks.]{\label{fig:ablation}
\includegraphics[width=5.5cm]{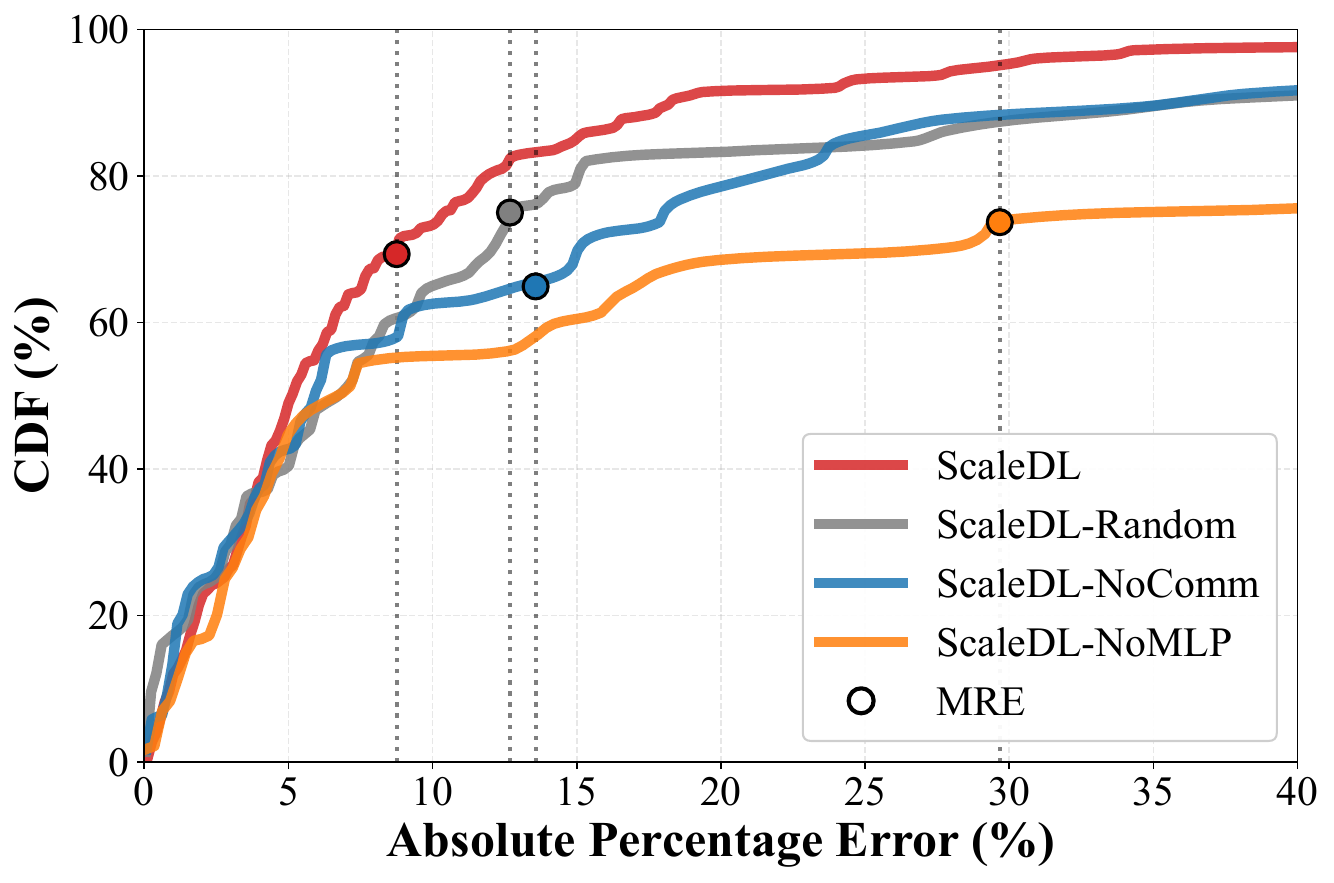}
}
\hspace{-5mm}
\caption{Ablation study.}
\vspace{-0.4cm}
\end{figure}

In this section, we performed different data sampling strategies and three ablation tests to assess the impact of different components in ScaleDL.


\textbf{Data-efficient Training.} As shown in Figure~\ref{fig:samples}, when achieving similar performance to the Full strategy (which trains on the entire dataset), D-optimal sampling requires only one-fifth of the samples compared to the Full strategy. 
At the same time, as illustrated in the Figure~\ref{fig:ablation}, ScaleDL-Random replaces D-optimal sampling with random sampling under the same data collection budget, resulting in a 1.4$\times$ higher MRE value. 
The above results demonstrates that D-optimal sampling reduces profiling overhead while maintaining accuracy.

\textbf{Component Contributions}. We assessed the impact of individual components in ScaleDL. In ScaleDL-NoComm, removing the communication model prevents ScaleDL from explicitly capturing the communication overhead. In ScaleDL-NoMLP, omitting the global encoder causes the GNN model to rely solely on local information integration from nodes and edges. These tests highlight the significance of each component: the MRE of ScaleDL-NoComm is 1.5 $\times$ that of ScaleDL, indicating the importance of communication modeling. Meanwhile, the largest error increase occurs in ScaleDL-NoMLP, where the MRE is 3.3$\times$ higher, showing that removing the MLP branch limits the model’s ability to handle global information and reduces the effectiveness of cross-layer interaction modeling.

\section{Conclusion}
This paper introduces ScaleDL, a runtime prediction framework for distributed DNN workloads. ScaleDL enhances prediction accuracy and generalizability  by integrating layer-wise computation model with the cross-layer interaction and communication model in the computation graph.
ScaleDL outperforms baseline models in both ID and OOD scenarios, reducing MRE and RMSE up to 6$\times$ and 5$\times$. The ablation study also demonstrates the effectiveness and significance of ScaleDL's design choices.

\bibliographystyle{IEEEtran}
\bibliography{reference}

\end{document}